\documentclass[conference]{IEEEtran}
\usepackage{cite}
\usepackage{amsmath,amssymb,amsfonts}
\usepackage{algorithmic}
\usepackage{graphicx}
\usepackage{textcomp}
\usepackage{xcolor}
\usepackage[english]{babel}
\usepackage[utf8]{inputenc}
\usepackage{csquotes}

\def\BibTeX{{\rm B\kern-.05em{\sc i\kern-.025em b}\kern-.08em
    T\kern-.1667em\lower.7ex\hbox{E}\kern-.125emX}}
\begin{document}

\title{Deep Visual Odometry Methods for Mobile Robots}
\author{Jahanzaib Shabbir and Thomas Kruezer}
\date{31/07/2018}

\maketitle

\begin{abstract}
Technology has made navigation in 3D real time possible and this has made possible what seemed impossible. This paper explores the aspect of deep visual odometry methods for mobile robots. Visual odometry has been instrumental in making this navigation successful. Noticeable challenges in mobile robots including the inability to attain Simultaneous Localization and Mapping have been solved by visual odometry through its cameras which are suitable for human environments. More intuitive, precise and accurate detection have been made possible by visual odometry in mobile robots. Another challenge in the mobile robot world is the 3D map reconstruction for exploration. A dense map in mobile robots can facilitate for localization and more accurate findings.   
\end{abstract}

\section{Visual odometry in mobile robots}
Mobile robot applications heavily rely on the ability of the vehicle to achieve accurate localization. It is essential that a robot is able to maintain knowledge about its position at all times in order to achieve autonomous navigation. To attain this, various techniques, systems and sensors have been established to aid with mobile robot positioning including visual odometry \cite{turan2017non}. Importantly, the adoption of Deep Learning based techniques was inspired by the precision to find solutions to numerous standard computer vision problems including object detection, image classification and segmentation. Visual odometry involves the pose estimation process that involves a robot and how they use a stream of images obtained from cameras that are attached to them \cite{li2017undeepvo}. The main aim of visual odometry is the estimations from camera pose. It is an approach that avoids contact with the robot for the purpose of ensuring that the mobile robots are effectively positioned. For this reason, the process is quite a challenging task that is related to mapping and simultaneous localization whose main aim is to generate the road map from a stream of visual data \cite{wang2017deepvo}. Estimates of motion from pixel differences and features between frames are made based on cameras that are strategically positioned. 
	For mobile robots to achieve an actively controlled navigation, a real time 3D and reliable localization and reconstruction of functions is an essential prerequisite \cite{turan2017endo}. Mobile robots have to perform localization and mapping functions simultaneously and this poses a major challenge for them. The Simultaneous Localization and Mapping (SLAM) problem has attracted attention as various studies extensively evaluate it \cite{mohanty2016deepvo}. To solve the SLAM problem, visual odometry has been suggested especially because cameras provide high quality information at a low cost from the sensors that are conducive for human environments \cite{Turan2017SixDL}. The major advances in computer vision also make possible quite a number of synergistic capabilities including terrain and scene classification, object detection and recognition. 
	Notably, the visual odometry in mobile robot have enabled for more precise, intuitive and accurate detection. Although there has been significant progress in the last decade to bring improvements to passive mobile robots into controllable robots that are active, there are still notable challenges in the effort to achieve this. Particularly, a 3D map reconstruction that is fully dense to facilitate for exploration still remains an unsolved problem. It is only through a dense map that mobile robots can be able to more reliably do localization and ultimately leading to findings that are more accurate \cite{konda2015learning} \cite{TuranAKS17}. According to Turan (\cite{turan2018sparse}), it is essential that adoptions of a comprehensive reconstruction on the suitable 3D method for mobile robots be adopted. This can be made possible through the building of a modular fashion including key frame selection, pre-processing, estimates on sparse then dense alignment based pose, shading based 3D and bundle fusion reconstruction \cite{turan2018magnetic}. There is also the challenge of the real time precise localization of the mobile robots that are actively controlled. The study by \cite{turan2018deep}, which employed quantitative and qualitative in trajectory estimations sought to find solution to the challenge of precise localization for the endoscopic robot capsule. The data set was general and this was ensured through the fitting of 3 endoscopic cameras in different locations for the purpose of capturing the endoscopic videos \cite{abs-1803-01047}. Stomach videos were recorded for 15 minutes and they contained more than 10,000 frames. Through this, the ground truth was served for the 3D reconstruction module maps’ quantitative evaluations \cite{TuranPJAKS17}. Its findings proposed that the direct SLAM be implemented on a map fusion based method that is non rigid for the mobile robots \cite{Turan2017}. Through this method, high accuracy is likely to be achieved for extensive evaluations and conclusions \cite{abs-1709-03401}. 
	The industry of mobile robots continues to face numerous challenges majorly because of enabling technology, including perception, artificial intelligence and power sources \cite{muller2017flowdometry}. Evidently, motors, actuators and gears are essential to the robotic world today. Work is still in progress in the development of soft robotics, artificial muscles and strategies of assembly that are aimed at developing the autonomous robot’s generation in the coming future that are power efficient and multifunctional. There is also the aspect of robots lacing synchrony, calibration and symmetry which serves to increase the photometric error. This challenge maybe addressed by adopting the direct odometry method \cite{yang2017direct}. Direct sparse odometry has been recommended by various studies since it has been found to reduce the photometric error. This can be associated to the fact that it combines a probabilistic model with joint optimization of model parameters \cite{turan2018sparse}. It has also been found to maintain high levels of consistency especially because it incorporates geometry parameters which also increase accuracy levels \cite{davison2003real}.
\section{Conclusion}
Visual odometry technology has made possible what seemed like impossible. Visual odometry has enabled mobile robots to have real time localization and mapping simultaneously thus eliminating to a great extent the SLAM problem. However, mobile robot technology has its own challenges as a result of shortcomings in the enabling technology space, including perception, artificial intelligence and power sources. This results to the inability of mobile robots to ensure accuracy since it lacks basic parameters including calibration and asymmetry which lead to increased photometric error. To counter this, direct odometry can come in handy since it incorporates calibrations. Nonetheless, there is notable progress in deep visual odometry methods for mobile robots in comparison to the past decades.

\end{document}